\documentclass[letterpaper, 10pt, conference]{ieeeconf}

\IEEEoverridecommandlockouts  
\usepackage{amsmath}
\usepackage{amssymb}
\usepackage{dsfont}
\usepackage{booktabs}
\usepackage{graphicx}
\graphicspath{{figures/}}
\usepackage{cite}
\usepackage[hidelinks]{hyperref}
\usepackage{flushend}
\usepackage[stretch=30]{microtype}

\title{\LARGE \bf
  DistAL: Distance-based Advantage Learning for VLA Fine-Tuning
}

\author{Reece O'Mahoney$^{1}$ and Ioannis Havoutis$^{1}$
  \thanks{$^{1}$The authors are with the Oxford Robotics Institute,
    University of Oxford, UK.
  {\tt\small \{reeceo,ioannis\}@robots.ox.ac.uk}}%
}

\begin{document}
\raggedbottom

\maketitle
\thispagestyle{empty}
\pagestyle{empty}


\begin{abstract}
  Vision-language-action models (VLAs) have transformed the field of
  robotic manipulation in recent years by combining the semantic
  understanding of LLMs with the precise control of flow-matching
  policies. Advantage conditioning is a recent technique that
  iteratively improves VLAs by training a value function on
  deployment data and using this to train an advantage-conditioned
  policy. Previous works have only applied simple, low-information
  success/failure rewards, which leave the value function unable to
  distinguish states of differing quality beyond how far along the
  task they appear. Motivated by an exploration of
  out-of-distribution (OOD) detection methods, we introduce
  Distance-based Advantage Learning (DistAL), which, by using an
  embedding space distance as a reward, produces a more informative
  value function and subsequently a higher downstream task success
  rate. We validate our method on a series of simulation benchmarks
  and dexterous bi-manual manipulation tasks on real hardware.
\end{abstract}


\section{Introduction}
VLAs \cite{brohan2023rt2,kim2024openvla,black2024pi0,intelligence2025pi05}
have allowed for a level of semantic generalisation and
dexterous manipulation not possible with previous generations of
robotic control techniques. In particular, the rise of flow-matching
\cite{lipman2023flow,black2024pi0,intelligence2025pi05} experts for
generating continuous actions, as opposed to earlier models which
discretised actions into tokens~\cite{brohan2022rt1}, typically
predicted auto-regressively~\cite{brohan2023rt2,kim2024openvla},
has greatly enhanced
the ability of transformer-based policies to follow complex action
distributions accurately without drifting out of distribution.
However, given the uncertainty of the dynamics in contact-rich,
dexterous manipulation tasks, the policies can still fail to achieve
the kind of success rates that would make them viable for real-world
deployment. In response to this, the robotics community has been
actively developing strategies for improving robotic performance from
their own deployment data \cite{luo2024hilserl,intelligence2025pistar06}.

One popular and successful approach has been via the use of intervention data
\cite{luo2024hilserl,liu2022sirius,liu2024siriusfleet},
where a human operator takes over control
from the policy when, or preferably just before, it reaches some
failure state. While this has proven successful, it significantly
increases the hardware and operational complexity needed to
develop a reliable model, as well as requiring an expert
teleoperator to be ready to intervene on potentially dozens of hours
of trials. Using online reinforcement learning to fine-tune VLAs,
either directly
\cite{mark2024parl,chen2025conrft,guo2025irevla,tan2025riptvla,lu2025vlarl}
or by distilling RL specialists into them \cite{xu2024rldg}, has
been another popular direction,
but suffers from issues of poor sample efficiency, instability, and
extremely high computational cost given the large parameter counts
of the base models. Recent work has improved sample efficiency
\cite{xu2026rltoken}, but still depends on human-provided rewards
and interventions during training.

Recently, advantage conditioning
\cite{frans2025diffusionguidance,intelligence2025pistar06} has
emerged as a promising new direction for leveraging deployment data.
It works by training a value function on the data, which is then used
to compute the advantage at each timestep. By thresholding a certain
advantage level as ``positive'', and optionally sharpening the
conditioning with classifier-free guidance
\cite{ho2022cfg}, this method is essentially offline RL. This does not require
an expert human teleoperator like intervention-based methods, and
also does not suffer from the same instability and sample efficiency
issues as online RL.

\begin{figure*}[t]
  \centering
  \includegraphics[width=0.95\textwidth]{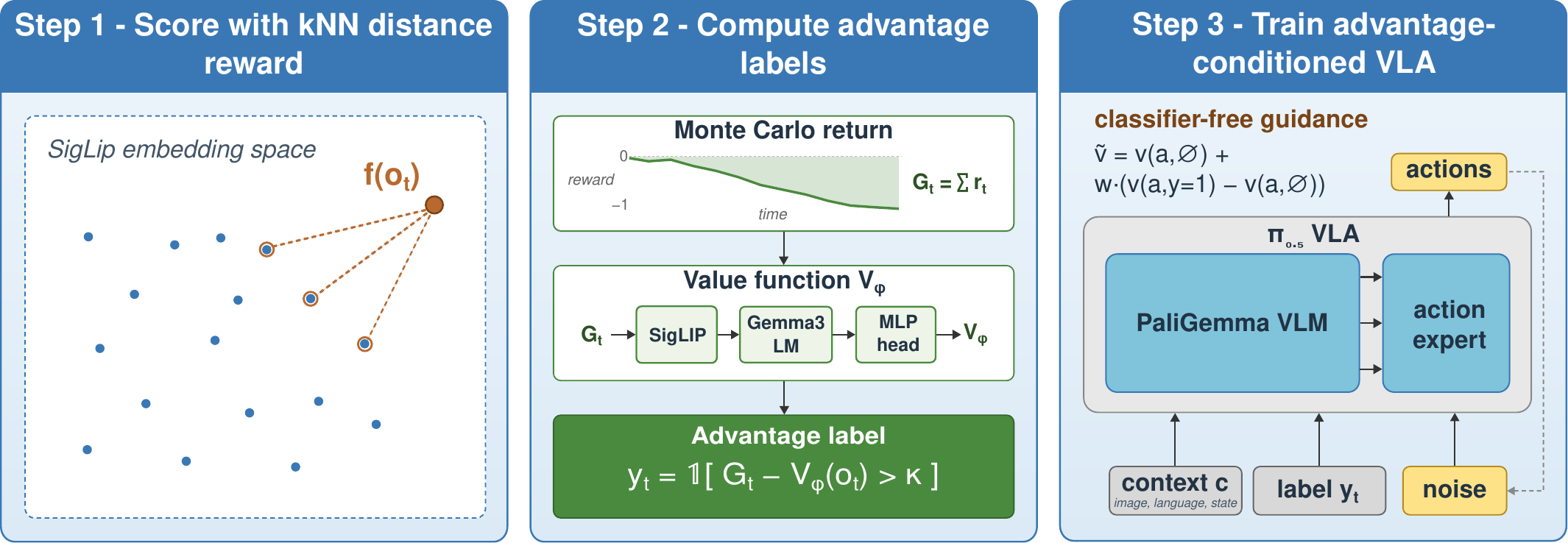}
  \caption{Overview of DistAL. Instead of the sparse success/failure
    reward used by prior advantage-conditioning methods, we score
    each deployment observation by its $k$-nearest-neighbours
    distance to the base VLA's training distribution in SigLIP
    feature space. The resulting dense reward yields a more
    informative value function and, in turn, a stronger
  advantage-conditioned policy.}
  \label{fig:overview}
\end{figure*}

Previous methods have only applied this technique to simple
success/failure rewards \cite{intelligence2025pistar06}, where a
constant per-step penalty means the only factor that raises a state's
value is the estimated number of steps remaining. States are therefore ranked
purely by how far along the task they appear, with no regard for any
other signal in the observations. For instance, an unstable grasp
that barely holds an object and a secure one receive the same value
despite very different odds of eventual success. This relationship is
somewhat captured by the failure penalty, but the credit assignment
is noisy and indirect. Custom, task-specific reward functions could
be used instead, but this comes with a substantial cost to
scalability, as the form and weighting of these rewards require
laborious tuning and/or manual frame-by-frame labelling.

We present \textit{Distance-based Advantage Learning} (DistAL) as an
alternative to both of these. DistAL is a drop-in replacement for
the reward signal: it slots directly into the advantage-conditioning
pipeline of \cite{intelligence2025pistar06} with the policy, value
function and training procedure left unchanged, isolating reward
design as the single variable in our comparisons. Our
approach is motivated by the idea that failure trajectories will be
OOD when compared to a reference dataset of successes. By exploring
the performance of different OOD detection methods at predicting
failure, we found that a \textit{kNN distance in image
embedding space} was most effective, both at the trajectory level and
at the per-timestep level that a value function actually consumes,
and thus use it as a reward function. Compared to a binary success
label, it reflects the \textit{immediate} quality of each observation
rather than relying on the failure being propagated backwards in time
through the value function. Our experimental evaluation shows that
downstream policies trained with this value function for advantage
conditioning exhibit improved performance on a range of simulated
and real tasks, with the largest gains where success and failure are
visually separable.

In summary, our contributions are:
\begin{enumerate}
  \item A comparison of out-of-distribution detection methods as
    predictors of VLA task failure, at both the trajectory and
    per-timestep level, identifying kNN distance in the base VLA's
    SigLIP feature space as the strongest signal.
  \item A controlled study of using this distance as a dense
    reward for advantage conditioning, holding the rest of the
    pipeline fixed, including an analysis of when the underlying
    assumption holds and when it does not.
  \item Validation on the LIBERO and LIBERO-plus simulated benchmark
    suites and on two dexterous bi-manual manipulation tasks on real
    robot hardware.
\end{enumerate}


\section{Related Work}
\label{sec:related-work}

\textbf{Vision-language-action models.} VLAs extend pre-trained
vision-language backbones with action prediction heads, inheriting the
semantic priors of internet-scale pre-training. Early models such as
RT-1 \cite{brohan2022rt1} discretised actions into tokens, and RT-2
\cite{brohan2023rt2} and OpenVLA \cite{kim2024openvla} predict these
tokens auto-regressively, which is too slow for high-frequency
control and performs poorly on bimanual tasks; OpenVLA-OFT
\cite{kim2025openvlaoft} mitigates this with parallel decoding. The
dominant approach more recently has been to generate continuous
action chunks with a diffusion or flow-matching head
\cite{chi2023diffusionpolicy,octo2024,liu2024rdt}, increasingly
coupled with a pre-trained VLM backbone
\cite{black2024pi0,intelligence2025pi05,nvidia2025groot,li2024cogact}.
These policies remain imitation-learned from demonstrations and cannot
improve from their own interactions, motivating the work below on
learning from deployment data.

\textbf{Improving VLAs from deployment data.} A first family exploits
human interventions during deployment: HIL-SERL \cite{luo2024hilserl}
combines interventions with online RL, while Sirius \cite{liu2022sirius}
and Sirius-Fleet \cite{liu2024siriusfleet} re-weight or solicit
interventions using trust and anomaly signals, requiring an expert
teleoperator throughout collection. A second family applies online RL to VLAs
\cite{mark2024parl,chen2025conrft,guo2025irevla,tan2025riptvla,lu2025vlarl},
or distils RL-trained specialists into them \cite{xu2024rldg},
contending with sample inefficiency, instability, and substantial
compute cost. A third family, the one DistAL belongs to, uses
offline value learning on logged trajectories. Building on offline-RL
foundations
\cite{kumar2020cql,kostrikov2021iql,peng2019awr,chen2021decisiontransformer},
recent works train a value function on logged data and use this
to improve the VLA. \emph{Steering Your Generalists}
\cite{nakamoto2024steering} re-ranks candidate actions by
value, and $\pi^{*}_{0.6}$/RECAP \cite{intelligence2025pistar06}
conditions a flow-matching VLA on a binary advantage label, optionally
sharpened with classifier-free guidance
\cite{ho2022cfg}. Both rely on sparse success/failure rewards that
make credit assignment more difficult. DistAL targets this weakness
by replacing the binary signal with a dense, embedding-distance
reward.

\textbf{Failure and out-of-distribution detection for policies.} A
separate literature recognises when a learned policy is failing or
operating out of distribution: anomaly scores in LLM-embedding space
\cite{sinha2024anomaly}, temporal-consistency and VLM progress checks
\cite{agia2024sentinel}, value functions reused as plan-feasibility
scores \cite{agia2022stap}, label-free uncertainty signals
\cite{xu2025faildetect}, and calibrated planner uncertainty
\cite{ren2023knowno}. These share our observation that learned
representations carry strong signal about eventual success.
Methodologically, our reward draws on the OOD-detection community:
feature-space nearest-neighbour distance \cite{sun2022knnood} and
class-conditional Mahalanobis distance \cite{lee2018mahalanobis} score
test points by distance to the training distribution, with deep
ensembles \cite{lakshminarayanan2017ensembles} as a generic uncertainty
backbone. DistAL borrows this distance-to-training-set signal but,
rather than using it for runtime rejection, repurposes it as a
per-timestep reward for offline value learning.


\section{Preliminaries}
\label{sec:preliminaries}

\textbf{Flow-matching VLAs.} VLAs predict a chunk of $H$ continuous
actions $a = (a^1, \ldots, a^H) \in \mathbb{R}^{H \times d_a}$ from an
observation--language context $c = (o, \ell)$ using a conditional
flow-matching head \cite{black2024pi0,lipman2023flow}. The head
learns a time-dependent velocity field $v_\theta(a_\tau, \tau, c)$
that transports a Gaussian noise sample $a_0 \sim \mathcal{N}(0, I)$
to a data sample $a_1$ along the linear interpolation $a_\tau =
(1-\tau)\, a_0 + \tau\, a_1$ for flow time $\tau \in [0,1]$. The
training objective is
\begin{equation}
  \mathcal{L}_{\text{FM}}(\theta) = \mathbb{E}_{\tau,\, a_0,\, (a_1,
  c) \sim \mathcal{D}} \Big[ \lVert v_\theta(a_\tau, \tau, c) - (a_1
  - a_0) \rVert^2 \Big].
  \label{eq:flow-matching}
\end{equation}
At inference an action chunk is produced by integrating $\dot a =
v_\theta(a, \tau, c)$ from $a_0 \sim \mathcal{N}(0, I)$ over $\tau
\in [0,1]$ with a small number of Euler steps, typically between 5 and 10.

\textbf{Advantage conditioning via classifier-free guidance.} Given a
dataset of deployment trajectories $\mathcal{D}_{deploy} = \{(o_t, a_t,
r_t)\}_{t=0}^{T}$ annotated with per-timestep rewards $r_t$,
advantage conditioning
\cite{intelligence2025pistar06} first fits a
state-value function $V_\phi$ by Monte Carlo regression onto the
empirical return (the original uses an $N$-step bootstrapped
estimate of the advantage; we use the full-episode return),
\begin{equation}
  \mathcal{L}_V(\phi) = \mathbb{E}_{(o_t, G_t) \sim \mathcal{D}}
  \Big[ \big( V_\phi(o_t) - G_t \big)^2 \Big], \;
  G_t = \sum_{k=0}^{T-t} r_{t+k},
  \label{eq:value-loss}
\end{equation}
and converts the resulting Monte Carlo advantage $\hat A_t = G_t -
V_\phi(o_t)$ into a conditioning label $y_t = \mathds{1}[\hat A_t >
\kappa]$ via a threshold $\kappa$. The VLA is then fine-tuned with
classifier-free guidance \cite{ho2022cfg} by dropping the advantage
label with probability $p$, which we represent as a null token $\varnothing$. At
inference, the velocity field is steered toward positive advantage actions via
\begin{equation}
  \begin{split}
    \tilde v_\theta(a_\tau, \tau, c, y{=}1) = {} & v_\theta(a_\tau, \tau, c,
    \varnothing) \\
    & + w \cdot \big( v_\theta(a_\tau, \tau, c, y{=}1) \\
    & \qquad\;\; - v_\theta(a_\tau, \tau, c, \varnothing) \big),
  \end{split}
  \label{eq:cfg}
\end{equation}
with guidance weight $w > 1$.


\section{Method}
\label{sec:method}

\begin{figure}[t]
  \centering
  \includegraphics[width=\linewidth]{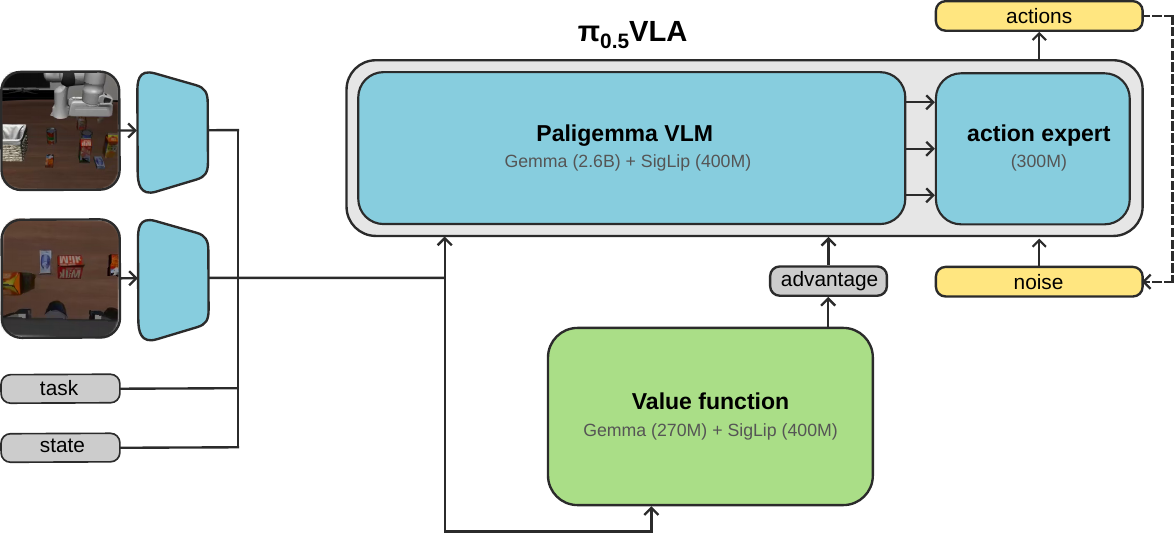}
  \caption{DistAL architecture from Section~\ref{ssec:architecture}.
    The value function
    $V_\phi$ is trained on deployment data using the kNN
    feature-distance reward, and the resulting advantage label is
  injected into the VLA via classifier-free guidance.}
  \label{fig:arch}
\end{figure}

The advantage-conditioning pipeline of Section~\ref{sec:preliminaries}
has so far been paired only with a binary success/failure reward
\cite{intelligence2025pistar06}, which, as noted above, ranks states
by estimated steps remaining and chance of success. DistAL replaces
this constant per-step penalty with a dense reward based on a
$k$-nearest-neighbours distance in the vision embedding space. We
describe the score calculation and encoder choice
(Section~\ref{ssec:knn}) and how the score plugs into the
rest of the pipeline (Section~\ref{ssec:training}).

\subsection{kNN feature-distance reward}
\label{ssec:knn}

Let $f: \mathcal{O} \to \mathbb{R}^{d_f}$ be a frozen image encoder
and let $\mathcal{I} = \{f(o^{(i)})\}_{i=1}^{N}$ denote an index of
embeddings computed over the observations in the dataset
$\mathcal{D}_{\text{train}}$ used to train the base VLA. For any
observation $o$ encountered at deployment, we define its
$k$-nearest-neighbours distance to the training distribution as
\begin{equation}
  s_k(o) = \frac{1}{k} \sum_{j \in \mathcal{N}_k(f(o);\,
  \mathcal{I})} \lVert f(o) - z_j \rVert_2,
  \label{eq:knn-score}
\end{equation}
where $\mathcal{N}_k(z; \mathcal{I})$ returns the indices of the $k$
nearest neighbours of $z$ in $\mathcal{I}$ under the Euclidean norm
and $z_j \in \mathcal{I}$ is the $j$th indexed embedding.
Observations close to the support of the training distribution
receive low scores; observations that drift away from it receive high
scores.

We then take the per-timestep reward used to drive the value function
in \eqref{eq:value-loss} to be the negative distance,
\begin{equation}
  r_t = -s_k(o_t)
  \label{eq:reward}
\end{equation}
normalised to $[-1, 0]$. This reward encodes the assumption that
observations close to the base VLA's training distribution are
better. That assumption is not universally true: a policy could in
principle raise its per-step reward by remaining in familiar early
states rather than progressing, and a valid recovery manoeuvre may
leave the training manifold even though it leads to success. We
therefore do not use the distance as a standalone objective. We
retain the large terminal penalty on failed episodes from
\cite{intelligence2025pistar06}, so that an episode which stays in
distribution but never completes the task still receives a low
return. The dense kNN distance supplies a per-step signal
of how far the policy has drifted from the training distribution,
while the large failure penalty preserves a strong signal of eventual
task success. Together they give the value function a measure both of
per-state quality and of final outcome. Section~\ref{ssec:hw-experiments}
shows the consequence of this assumption in practice: the method helps
most where failure is visually distinct from success, and least where it
is not. In Section~\ref{sec:result} we compare this metric with other
common OOD detection methods and show that it best predicts policy
failure, motivating its use as a reward function for advantage
learning. We found that task success was insensitive to the choice of
$k$, so we used $k=10$ as a balanced value.

The choice of encoder $f$ determines the space in which this distance
is calculated. In Section~\ref{sec:result} we also compare different
alternatives and find that a frozen
SigLIP~\cite{zhai2023siglip} embedding produced the best results. It
is important to note that this is not a base SigLIP model, but the
same fine-tuned vision encoder as the base VLA.

\subsection{Architecture}
\label{ssec:architecture}

Our overall architecture is shown in Fig.~\ref{fig:arch}. It is
largely similar to \cite{intelligence2025pistar06}, with modifications
or additions where details were unspecified in the original paper.

\textbf{Value function.} The value function $V_\phi$ consists of a
pre-trained Gemma3 \cite{gemmateam2025gemma3,google2025gemma3_270m} 270M
parameter language model and a
400M parameter SigLIP vision encoder. To predict values, we pass in
the same image, language and state conditioning $c$ that the policy
observes, and then pass the final output token through a 2-layer MLP
head. Instead of predicting a continuous target, we follow
\cite{farebrother2024stopregressing} and predict a categorical distribution over
discretised return bins, and train this with a cross-entropy loss
with the targets from \eqref{eq:value-loss}. It is worth noting
that we also tried the more complex HL-Gauss loss from
\cite{farebrother2024stopregressing} but found it provided no benefit
over cross-entropy.

\textbf{Advantage-conditioned policy.} Our VLA is built from a
fine-tuned $\pi_{0.5}$ \cite{intelligence2025pi05} model, which,
following $\pi_0$ \cite{black2024pi0}, uses
the PaliGemma \cite{beyer2024paligemma} 3B parameter VLM along with a
300M parameter flow-matching action expert. We inject the binary
advantage label into the VLM backbone by simply appending the text
``Advantage: positive'' or ``Advantage: negative'' to the end of the
tokenised language input depending on the threshold value. We also
follow $\pi^{*}_{0.6}$ and set this threshold so that 30\% of the
advantage labels are positive. In the unconditional case, nothing is
appended. We obtained the best results by fine-tuning only
the action expert and freezing the backbone. At inference the policy
is then conditioned with the positive advantage label, optionally
with guidance using \eqref{eq:cfg}.

\subsection{Training}
\label{ssec:training}

Training DistAL given a deployment dataset $\mathcal{D}_{deploy} = \{(o_t,
a_t)\}_{t=0}^{T}$ proceeds in three steps:
\begin{enumerate}
  \item \textbf{Score} every deployment observation with
    \eqref{eq:knn-score} against the training-set index
    $\mathcal{I}$, and assign the per-step reward $r_t = -s_k(o_t)$.
  \item \textbf{Fit} the value function $V_\phi$ onto the empirical
    return $G_t = \sum_{k=0}^{T-t} r_{t+k}$
    \eqref{eq:value-loss}, then label each timestep
    with $y_t = \mathds{1}[\hat A_t > \kappa]$ where $\hat A_t = G_t
    - V_\phi(o_t)$.
  \item \textbf{Fine-tune} the VLA with the advantage
    label $y_t$, dropping it to $\varnothing$ with probability $p$ at
    training time. At inference the policy is steered toward
    high-advantage actions via the classifier-free guidance velocity
    of \eqref{eq:cfg}.
\end{enumerate}


\section{Experimental Results}
\label{sec:result}

\subsection{OOD detection as a failure predictor}
\label{ssec:ood-eval}

We first ask which metric best predicts task success in order to
motivate the choice of reward function and encoding method for
advantage conditioning.

\textbf{Setup.} We collect rollouts from the base $\pi_{0.5}$ policy
on LIBERO, LIBERO-plus, and our two bi-manual hardware tasks, each
labelled with a binary success/failure outcome. For every observation
$o_t$ we compute each method's per-step score, and aggregate it to a
trajectory-level score by taking the mean. Max aggregation was also
tested but was found to perform worse across the board. We report
AUROC scores against the success label, so a random score gives $0.5$
and a perfect detector gives $1.0$. Scores less than $0.5$ indicate the
classification is inverted. We report both per-trajectory and
per-timestep AUROC.

\textbf{Methods.} We compare five detectors chosen to disentangle
three axes: where in the VLA the features are tapped (row~1 vs.~2),
the distance metric (row~1 vs.~3), and generative/action-space
alternatives (rows~4--5):
\begin{enumerate}
  \item \textbf{kNN -- SigLIP features (pre-LM).} Equation~\eqref{eq:knn-score}
    with the SigLIP vision encoder features of the base VLA, taken
    \emph{before} the language model. Our proposed reward.
  \item \textbf{kNN -- image output tokens (post-LM).} Same kNN score,
    but computed on the image-token activations at the
    \emph{output} of the VLA language model, to test whether
    LM-contextualised features carry a stronger failure signal than
    the raw vision features.
  \item \textbf{Mahalanobis -- SigLIP features.} Class-conditional
    Mahalanobis distance \cite{lee2018mahalanobis} in the same
    SigLIP feature space as row~1, to ablate the distance metric.
  \item \textbf{VAE likelihood.} Negative log-likelihood under a VAE
    trained on the same SigLIP features over
    $\mathcal{D}_{\text{train}}$, as a generative-model alternative
    to nearest-neighbour scoring.
  \item \textbf{Action-chunk variance.} Per-step standard deviation
    across 16 flow-matching samples drawn from the policy for the
    same context $c$.
\end{enumerate}

\begin{table}[t]
  \caption{Per-trajectory AUROC of each OOD detector at predicting
  rollout failure. Higher is better; random is $0.5$.}
  \label{tab:ood-auroc}
  \centering
  \footnotesize
  \setlength{\tabcolsep}{2.5pt}
  \begin{tabular}{lccccc}
    \toprule
    Method & LIBERO & LIBERO+ & Pen lid & Ethernet & Mean \\
    \midrule
    kNN, SigLIP (ours) & \textbf{0.82} &
    \textbf{0.84} & 0.66 & \textbf{0.94} & \textbf{0.82} \\
    kNN, LM image tokens & 0.31 & 0.53 & 0.66 & 0.76 & 0.57 \\
    Mahalanobis, SigLIP & 0.64 & 0.75 & 0.59 & 0.86 & 0.71 \\
    VAE likelihood                         & 0.71 & 0.80 &
    \textbf{0.70} & \textbf{0.94} & 0.79 \\
    Action-chunk variance & 0.60 & 0.68 & 0.51 & 0.54 & 0.58 \\
    \bottomrule
  \end{tabular}
\end{table}

\begin{table}[t]
  \caption{Per-timestep AUROC of each detector.}
  \label{tab:ood-auroc-perstep}
  \centering
  \footnotesize
  \setlength{\tabcolsep}{2.5pt}
  \begin{tabular}{lccccc}
    \toprule
    Method & LIBERO & LIBERO+ & Pen lid & Ethernet & Mean \\
    \midrule
    kNN, SigLIP (ours) & \textbf{0.70} &
    \textbf{0.76} & 0.56 & \textbf{0.65} & \textbf{0.67} \\
    kNN, LM image tokens & 0.40 & 0.53 &
    0.56 & 0.56 & 0.51 \\
    Mahalanobis, SigLIP & 0.57 & 0.64 & 0.52 & 0.58 & 0.58 \\
    VAE likelihood                         & 0.64 & 0.73 &
    \textbf{0.59} & 0.64 & 0.65 \\
    Action-chunk variance & 0.52 & 0.56 & 0.45 & 0.49 & 0.51 \\
    \bottomrule
  \end{tabular}
\end{table}

\begin{figure*}[t]
  \centering
  \includegraphics[width=0.9\textwidth]{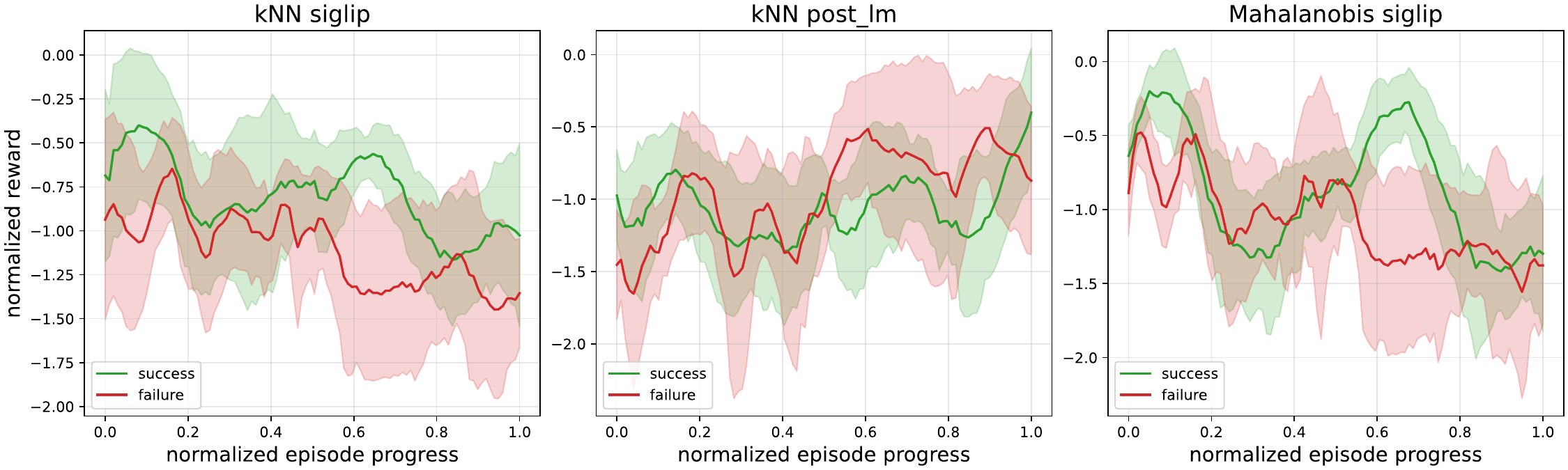}
  \caption{Per-step rewards along a successful and a failed rollout of
    the same task, averaged over 10 random episodes of each and
    min-max normalised to $[0,1]$ per method. The kNN score separates
    the two outcomes far more cleanly than the other two
    embedding-based detectors, which repeatedly cross over -- hence its
    much higher AUROC. All scores drift negatively as the episode
    progresses, likely because final states are more diverse than initial ones.
  }
  \label{fig:rollout-scores}
\end{figure*}

\begin{table*}[t]
  \caption{Success rate ($\%$) on LIBERO-10 and per-perturbation on
  LIBERO-plus.}
  \label{tab:sim-success}
  \centering
  \setlength{\tabcolsep}{6pt}
  \begin{tabular}{l c @{\hskip 1.5em} ccccccc c}
    \toprule
    & & \multicolumn{7}{c}{LIBERO-plus (per perturbation type)} & \\
    \cmidrule(lr){3-9}
    Method & LIBERO-10 & Layout & Camera & Light & Init. & Noise & Lang. &
    Texture & Mean \\
    \midrule
    Base $\pi_{0.5}$                & 94.8 & \textbf{77.8} & 41.5 &
    \textbf{96.0} &
    65.3 & 59.8 & \textbf{76.6} & 89.3 & 75.1 \\
    SFT (success-filtered)      & 94.4 & 77.5 & 41.2 & 95.8 &
    66.7 & 59.5 & 73.6 & 90.0 & 74.8 \\
    RECAP-style binary reward   & 95.0 & 64.0 & 61.9 & 95.6 &
    \textbf{67.3} & \textbf{76.1} & 65.2 & 92.3 & 77.2 \\
    \midrule
    DistAL (ours)               & \textbf{96.2} & 64.0   &
    \textbf{66.7}   & 93.3
    & \textbf{67.3}   & 73.9   & 65.6   & \textbf{100.0}   & \textbf{78.4}   \\
    \bottomrule
  \end{tabular}
\end{table*}

\textbf{AUROC Results.} Table~\ref{tab:ood-auroc} reports
per-trajectory AUROC and Table~\ref{tab:ood-auroc-perstep} reports
per-timestep AUROC. The per-timestep score is the more relevant of
the two, since the value function is fitted to
per-step rewards and must assign credit locally rather than
classify whole episodes. Our kNN distance on the base VLA's
pre-LM SigLIP features is the strongest detector on both
measures ($0.82$ per-trajectory and $0.67$ per-timestep mean),
ahead of the VAE likelihood ($0.79$ and $0.65$), the only baseline
that stays competitive across all four settings. Mahalanobis distance
in the same feature space trails ($0.71$): the SigLIP space carries the failure
signal, but the kNN metric exploits it more effectively than a
class-conditional Gaussian fit. The post-LM image tokens are far less
reliable ($0.57$ mean, even inverting to $0.31$ on LIBERO), indicating
that language-model contextualisation washes out rather than sharpens
the signal. The pen-lid task is the hardest setting for every
detector: its failures involve a correct grasp that slips, which
is hard to detect visually. It is also the one place our method is
not top, trailing the VAE by $0.04$ ($0.66$ vs.\ $0.70$); the more separable
ethernet task lets our kNN distance tie for the best per-trajectory
score ($0.94$). Overall, our method retains the best mean on both measures. The
per-timestep AUROCs preserve the same ordering as the per-trajectory
scores at uniformly lower values, as expected from noisier per-frame
labels: an individual frame early in a rollout that eventually fails
is often indistinguishable from one in a rollout that succeeds. The
pen-lid task is again the weak point, with a per-timestep AUROC of
$0.56$ that is close to chance, and we return to this when
interpreting the hardware results.

Plotting the per-step scores along successful and failed rollouts
(Fig.~\ref{fig:rollout-scores})
illustrates why the kNN distance wins, with a much cleaner separation
between successful and failed rollouts than the other detectors. This
figure is also the most direct evidence that the signal is usable for
local credit assignment: the gap between the two curves opens
progressively through the episode rather than only at the terminal
frame, so the value function receives a graded, timed signal rather
than a delayed one. This is the empirical motivation for using the
kNN feature distance as the reward in Section~\ref{ssec:knn}; the
remaining experiments in this section evaluate the downstream
advantage-conditioned policy that results.

\begin{figure*}[t]
  \centering
  \includegraphics[width=0.9\textwidth]{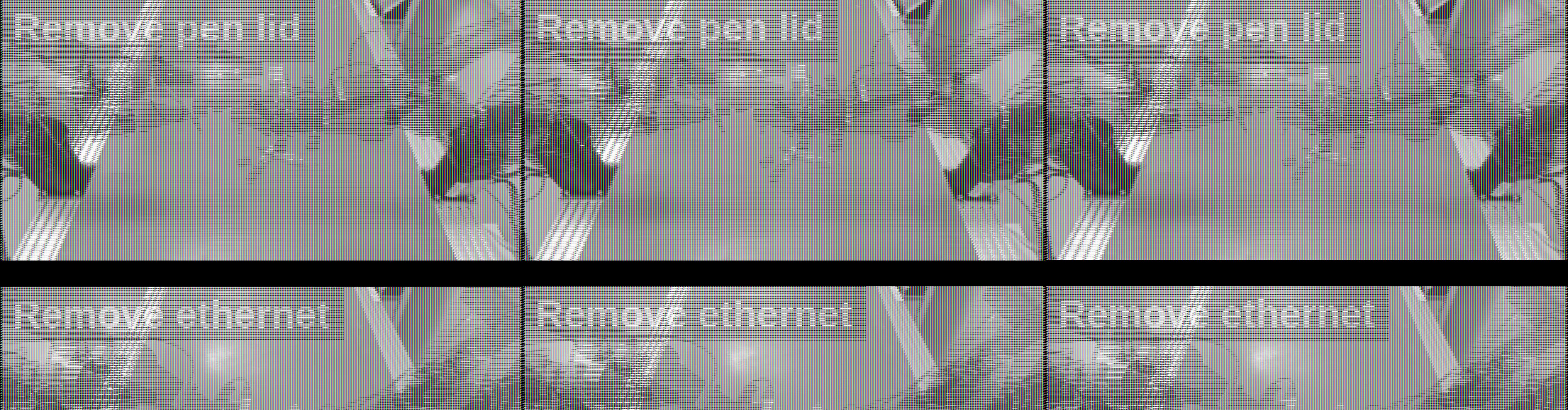}
  \caption{Keyframes of the two bi-manual Piper tasks: \emph{remove
    pen lid} (top) and \emph{remove ethernet cable} (bottom).
  }
  \label{fig:hw-tasks}
\end{figure*}

\subsection{Simulation experiments}
\label{ssec:sim-experiments}

Having established that the kNN feature distance is a strong
predictor of failure, we now evaluate whether using it as the reward
function for advantage conditioning translates into improved
downstream task performance.

\textbf{Setup.} We evaluate on LIBERO-10 (also called LIBERO-Long, the
ten long-horizon tasks of LIBERO-100~\cite{liu2023libero}) and on
LIBERO-plus~\cite{fei2025liberoplus}. We omit the
LIBERO-Spatial, -Object, and -Goal suites as $\pi_{0.5}$ already
saturates them, leaving no headroom to measure improvements. For
LIBERO-plus we report per-\emph{perturbation type} rather than
per-suite, to reveal which robustness axes the advantage signal
actually improves. We start from the same fine-tuned $\pi_{0.5}$ base
policy and collect a fixed deployment dataset
$\mathcal{D}_{\text{deploy}}$ of 50
rollouts per task with success/failure labels. All
advantage-conditioning methods are trained on the same dataset
to isolate the effect of the reward function. We report mean success
rate over 3 seeds, each with 50 episodes per task.

\textbf{Baselines.} We compare DistAL against three points of comparison
chosen to isolate the contribution of the dense distance reward:
\begin{enumerate}
  \item \textbf{Base $\pi_{0.5}$.} The fine-tuned base policy used to
    collect $\mathcal{D}_{\text{deploy}}$, without any advantage conditioning.
  \item \textbf{SFT (success-filtered).} The base policy further
    fine-tuned on only the successful trajectories of
    $\mathcal{D}_{\text{deploy}}$.
  \item \textbf{RECAP-style binary reward~\cite{intelligence2025pistar06}.}
    The same pipeline as DistAL but with the sparse success/failure
    reward and per-step penalty used by prior work.
\end{enumerate}

\textbf{Results.} Table~\ref{tab:sim-success} reports success rates
per task on LIBERO-10 and per perturbation type on LIBERO-plus.
DistAL is the strongest method on every aggregate: it
achieves the best LIBERO-10 score ($96.2\%$) and the highest overall
mean across LIBERO-10 and LIBERO-plus ($78.4\%$, $+3.3$ over base).
Most of that gain comes from advantage conditioning itself, which the
binary-reward baseline shares; the dense distance reward adds a
further $+1.2$ on both aggregates. This margin is consistent in sign
but small relative to seed-to-seed variation, so aggregate simulation
success alone does not separate the two rewards. Where they do
separate is on the individual perturbation types below and on
hardware, where the ethernet task shows the largest advantage of the
dense reward in the paper. SFT, by contrast, is indistinguishable
from the base policy ($-0.3$ on the overall mean): re-training on the
successful slice of $\mathcal{D}_{\text{deploy}}$ extracts no additional
signal, so the gains of the advantage-conditioned methods come from
the value function rather than from the data subset.

The per-perturbation breakdown reveals where each reward shape pays
off. Both advantage-conditioning methods deliver their largest gains
on the perturbations that the base policy handles worst:
\emph{Camera} ($+25.2$ for DistAL and $+20.4$ for the binary reward
over base), \emph{Noise} ($+14.1$ and $+16.3$), and \emph{Texture}
($+10.7$ and $+3.0$), likely because they incur the
largest visual shift from the training distribution and hence are
most suited to DistAL's reward function. Our approach also widens the margin
most clearly on \emph{Camera} and \emph{Texture}, where it reaches a perfect
$100.0\%$. The picture is not uniformly positive. On
\emph{Layout} and \emph{Language} both advantage-conditioned methods
regress substantially relative to base ($77.8 \to 64.0$ and
$76.6 \to 65.6$ for DistAL), and on \emph{Light} and \emph{Noise}
DistAL is $2.3$ and $2.2$ points below the binary reward. The
\emph{Layout} and \emph{Language} regressions are almost identical for
the two rewards ($64.0$ vs.\ $64.0$ and $65.6$ vs.\ $65.2$), which
indicates they are a property of advantage conditioning on this
deployment dataset rather than of the distance reward. We believe
the cause is that these perturbations leave the visual scene close
to the training distribution while changing the correct behaviour,
so neither reward can separate success from failure early, and the
conditioning trades in-distribution performance for recovery on
visually shifted observations. The \emph{Light} and \emph{Noise}
deficits are specific to DistAL and are within seed-to-seed
variation, but they are consistent with the distance reward
penalising benign visual shift that does not actually threaten the
task.

The comparison between DistAL and the binary-reward baseline isolates
the contribution of the dense distance signal. DistAL wins on $5$ of
the $8$ columns and matches on a sixth, with its largest individual
gains coming on \emph{Camera} ($+4.8$) and \emph{Texture} ($+7.7$),
and loses on the two noted above. The pattern, gains where the
perturbation moves the observation away from the training
distribution and losses where it does not, is what the reward's
assumption predicts. The next section evaluates whether the gains
transfer to real hardware.

\subsection{Hardware experiments}
\label{ssec:hw-experiments}

We now evaluate whether the gains observed in simulation transfer to
a real bi-manual platform, where contact dynamics and control are
substantially harder than in LIBERO.

\textbf{Setup.} We deploy on a bi-manual setup consisting of two
Piper arms, with a wrist camera on each arm and an external scene camera.
We evaluate on two contact-rich, dexterous tasks: \emph{remove pen
lid}, which requires one arm to stabilise the pen body while the
other grasps and pulls the lid clear; and \emph{remove ethernet
cable}, which requires one arm to brace the port while the other
grips and unplugs the cable (Fig.~\ref{fig:hw-tasks}). We collect 50
tele-operated demonstrations per task to fine-tune the base
$\pi_{0.5}$, then a deployment dataset $\mathcal{D}_{\text{deploy}}$ of 50 more
rollouts with success labels. As in simulation, all
advantage-conditioning methods are trained on the \emph{same}
deployment dataset to isolate the effect of the reward function. We
report mean success rate over 50 evaluation rollouts per task. We
compare DistAL against the two most informative baselines carried over
from the simulation study: the fine-tuned \textbf{base $\pi_{0.5}$}
without advantage conditioning, and the \textbf{RECAP-style reward}.

\begin{table}[t]
  \caption{Success rate ($\%$) on the two bi-manual Piper tasks,
  averaged over $50$ rollouts per task. Higher is better.}
  \label{tab:hw-success}
  \centering
  \setlength{\tabcolsep}{9pt}
  \begin{tabular}{lccc}
    \toprule
    Method & Pen lid & Ethernet & Mean \\
    \midrule
    Base $\pi_{0.5}$            & 42.0 & 46.0 & 44.0 \\
    RECAP-style binary reward   & 48.0 & 72.0 & 60.0 \\
    \midrule
    DistAL (ours)               & \textbf{52.0} & \textbf{87.0} &
    \textbf{69.5} \\
    \bottomrule
  \end{tabular}
\end{table}

\textbf{Results.} Table~\ref{tab:hw-success} reports success rates
across the two tasks. On \emph{remove ethernet cable}, DistAL reaches
$87\%$ against $46\%$ for the base policy and $72\%$ for the binary
reward, our strongest single result. On \emph{remove pen lid}, DistAL
is again the best of the three at $52\%$, a $10$-point gain over
base, but only $4$ points above the binary reward, a margin that is
not statistically significant at $50$ rollouts. We therefore treat the
two methods as tied on this task.

This split matches the failure-prediction results of
Section~\ref{ssec:ood-eval}. Ethernet failures (a cable left plugged
in, an arm that misses the port) are visually obvious and the kNN
distance separates them with a per-trajectory AUROC of $0.94$.
Pen-lid failures are a correctly formed grasp that slips under load,
which is a contact event the image does not show, and the per-timestep
AUROC there is $0.56$. When the reward cannot see the failure it
degrades to the binary reward plus noise, and the downstream policy
behaves accordingly. DistAL is therefore most useful when the
task's failure modes are visually distinguishable from its success
modes, and we would expect it to give little benefit on tasks whose
failures are primarily haptic.

Together with the simulation results, this supports DistAL as a
low-cost modification to sparse-reward advantage conditioning that
helps where its assumption holds, and identifies the visual
separability of failure as the condition to check before applying
it.


\section{Conclusion and Limitations}
\label{sec:conclusion}

We presented \emph{Distance-based Advantage Learning} (DistAL), a
modification to advantage-conditioned VLA fine-tuning that
replaces the sparse success/failure reward with a dense per-step
reward based on the $k$-nearest-neighbours distance to the base VLA's
training distribution in SigLIP feature space, keeping the rest of
the pipeline fixed. We first showed that this distance is the
strongest predictor of failure among a range of OOD-detection
baselines at both the trajectory and per-timestep level, motivating
its use as a reward function. We then showed that the resulting
advantage-conditioned policy improves over the base $\pi_{0.5}$
policy on LIBERO-10, LIBERO-plus, and two bi-manual manipulation
tasks on real hardware, and over the binary-reward baseline where
failure is visually separable from success, most clearly on the
\emph{Camera} and \emph{Texture} perturbations and on the ethernet
task. Where it is not, on the pen-lid task and on perturbations
that leave the scene visually unchanged, the two rewards are
indistinguishable at our sample sizes.

DistAL inherits the assumptions of advantage conditioning: it relies
on its own deployment data and cannot improve a policy until that policy
already attains the baseline success rate needed to produce a
meaningful signal, and what this level is remains an open question.
Two failure modes stem from the reward itself. First, the score is
purely image-based and ignores force, contact stability, robot state
and language, which limited performance on the \emph{remove pen lid}
task, where the
failure is a slip the camera cannot see. Second, because the reward
favours proximity to the base VLA's training distribution, a policy
can in principle raise its reward by staying in familiar states
rather than solving the task, and may be penalised for a novel but
successful recovery; we mitigate this with the terminal penalty and
by bounding the reward (Section~\ref{ssec:knn}) but have not
characterised it beyond the per-perturbation results in
Section~\ref{ssec:sim-experiments}. Our evaluation is also limited in
scope: a single base VLA ($\pi_{0.5}$), two real-world tasks, the
LIBERO suites, and three training seeds per configuration, so the
smaller differences we report are not statistically resolved and
generality across embodiments and task families remains to be
established. We have not ablated the terminal penalty, the reward
normalisation, or the downstream effect of the alternative OOD scores in
Table~\ref{tab:ood-auroc}. Promising future directions include those
ablations, incorporating force into the distance score, combining the
distance with an explicit task-progress signal, characterising the
minimum success rate required for improvement, and performing
multiple rounds of advantage conditioning to compare how the methods scale.

\section*{ACKNOWLEDGMENT}

Generative AI tools (Claude, Anthropic; ChatGPT, OpenAI) were used
throughout the preparation of this work, including drafting and
editing text in all sections, generating and debugging code for
experiments and figures, and assisting with the analysis of results.
All AI-generated content was reviewed and verified by the authors,
who take full responsibility for the final manuscript.


\bibliographystyle{IEEEtran}
\bibliography{main}

\end{document}